\documentclass[final,2p,times,twocolumn]{elsarticle}

\usepackage{prletters}
\usepackage{amssymb}

\usepackage{algorithm}
\usepackage{algorithmic}
\usepackage{hyperref}       
\usepackage{url}            
\usepackage[normalem]{ulem}
\usepackage{booktabs}       
\usepackage{amsfonts}       
\usepackage{nicefrac}       
\usepackage{microtype}      
\usepackage{xcolor}         

\usepackage{fixltx2e}
\usepackage{amsmath}

\usepackage{graphicx}
\usepackage{caption}
\usepackage{enumitem}
\usepackage{adjustbox}
\usepackage{pgfplots}
\usepackage{subcaption}
\usepackage{multirow}

\newcommand{\bfg}{\mathbf{g}}

\newcommand{\bfz}{\mathbf{z}}

\newcommand{\bfs}{\mathbf{s}}

\newcommand{\bfa}{\mathbf{a}}

\newcommand{\bff}{\mathbf{f}}
\newcommand{\bfy}{\mathbf{y}}
\newcommand{\bfK}{\mathbf{K}}

\newcommand{\bfA}{\mathbf{A}}
\newcommand{\bfD}{\mathbf{D}}

\newcommand{\bfS}{\mathbf{S}}

\newcommand{\bfP}{\mathbf{P}}

\journal{Pattern Recognition Letters}

\begin{document}

\begin{frontmatter}




\title{Global-Local Graph Neural Networks for Node-Classification}

\author[inst1,inst2]{Moshe Eliasof\corref{cor}}
\cortext[cor]{Corresponding author}
\ead{eliasof@post.bgu.ac.il}

\author[inst1]{Eran Treister}
\ead{erant@cs.bgu.ac.il}

\affiliation[inst1]{organization={Ben-Gurion University of the Negev, Department of Computer Science},
            city={Beer Sheva}, 
            country={Israel}}

\affiliation[inst2]{organization={University of Cambridge, Department of Applied Mathematics},
            city={Cambridge}, 
            country={United Kingdom}}

\begin{abstract}
The task of graph node classification is often approached \textcolor{black}{by utilizing} a \emph{local} Graph Neural Network (GNN), that learns only local information from the node input features and their adjacency. In this paper, we propose to \textcolor{black}{ improve the performance of node classification GNNs by utilizing both global and local information, specifically by } learning \emph{label}- and \emph{node}- features. We therefore call our method Global-Local-GNN (GLGNN).
To learn proper label features, for each label, we maximize the similarity between its features and nodes features that belong to the label, while maximizing the distance between nodes that do not belong to the considered label. We then use the learnt label features to predict the node classification map. We demonstrate our GLGNN using three different GNN backbones, and show that our approach improves baseline performance, revealing the importance of global information utilization \textcolor{black}{for node classification}.
\end{abstract}


\end{frontmatter}



\section{Introduction}
\label{sec:intro}
The field of Graph Neural Networks (GNNs) has gained large popularity in recent years~\cite{kipf2016semi} in a  variety of fields and applications such as computer graphics and vision~\cite{wang2018dynamic}, social network analysis~\cite{kipf2016semi} computational biology \cite{eliasof2021mimetic} and others. \textcolor{black}{The appeal of GNNs is their generalization of Convolutional Neural Networks (CNNs) \cite{KrizhevskySutskeverHinton2012, he2016deep} that operate on \textit{structured} information such as images or 1D sequences. That is, GNNs allow to also process unstructured data, such as point clouds, molecular structures, social networks, and more.} 
 Despite the success of GNNs, in the context of node classification, most of the methods consider only nodal (i.e., local) information by performing local neighbourhood aggregations and $1\times1$ convolutions, e.g., ~\cite{kipf2016semi,velickovic2018graph,chen20simple}. An interesting question is whether the incorporation of \emph{global} information is useful in the context of locally oriented tasks, such as node classification. \textcolor{black}{For example, it was shown in \cite{feng2022local} that using global information helps object detection - a locally oriented task. In the context of GNNs, this}  question was recently addressed using \textcolor{black}{the} Laplacian eigenvectors as positional encoding \cite{dwivedi2020benchmarkgnns}, learning global  \textcolor{black}{graph} features through convolutional and motif-based features \cite{li2021lgl} and input feature propagation \cite{huang2022from}. The common theme of those methods, that show accuracy improvement compared to using local features only, is that they directly rely on the graph connectivity to produce global features.
In contrast, in this paper we propose to utilize \textcolor{black}{node class} label (i.e., global) information with node (local) information to improve the training and performance of GNNs \textcolor{black}{on classification tasks}.  \textcolor{black}{Due to its \textbf{g}lobal and \textbf{l}ocal features utilization, we} call our method GLGNN.  In particular, we propose to learn a feature vector for each label (class) in the data, which is then used to determine the final prediction map and is mutually utilized with the learnt node features. Because our method is based on learning global features that scale as the number of labels in the dataset, our method does not add significant computational overhead compared to the backbone GNNs, \textcolor{black}{ and is suitable for classification tasks.} We show the generality of this approach by demonstrating it on GCN \cite{kipf2016semi}, GAT \cite{velickovic2018graph}, and GCNII \cite{chen20simple} on a variety of node-classification datasets, both in semi- and fully-supervised settings. Our experiments reveal that our GLGNN approach is beneficial for all considered datasets. Also, we illustrate the learnt global features compared to the node features for a qualitative assessment of our method. 
Our contributions are: 
(i) We propose to learn \emph{label} features to capture global information of the input graph. (ii) \textcolor{black}{We fuse label and node  features} to predict a node classification map.
(iii) We  illustrate the learnt label features in Figure \ref{fig:tsne} and quantitatively demonstrate the benefit of using GLGNN approach on \textcolor{black}{12} real-world datasets.

\section{Related work}
\label{sec:related}
\subsection{Graph Neural Networks}
Typically, graph neural networks (GNNs) are categorized into spectral~\cite{bruna2013spectral} 
and spatial~\cite{kipf2016semi,gilmer2017neural} 
types. While the former learns a global convolution kernel, it scales as the number of nodes in the graph $n$ and is of higher computational complexity. To obtain local convolutions, spatial GNNs formulate a local-aggregation scheme that is usually implemented using the Message-Passing Neural Network mechanism~\cite{gilmer2017neural}, where each node aggregates features (messages) from its neighbours, according to some policy.
In this work, we follow the latter, whilst adding a global mechanism by learning label features to improve \textcolor{black}{downstream} accuracy on node classification tasks. \textcolor{black}{In our experiments, we will show that our proposed mechanism is also beneficial in terms of accuracy for graph classification.}

\subsection{Improved training of GNNs}
To improve accuracy, recent works \textcolor{black}{have} introduce\textcolor{black}{d} new training policies, objective functions, and augmentations. A common trick for training on small datasets like Cora, Citeseer, and Pubmed is the incorporation of Dropout \cite{srivastava2014dropout} after every GNN layer, which has become a standard practice \cite{kipf2016semi,chen20simple}. Other methods suggest randomly alternating the data rather than the GNN neural units. For example,
DropEdge \cite{Rong2020DropEdge:} and DropNode \cite{do2021graph_dropnode} randomly drop graph edges and nodes, respectively. 
In the method PairNorm \cite{Zhao2020PairNorm:}, the authors propose a normalization layer that alleviates the over-smoothing phenomenon in GNNs \cite{measuringoversmoothing}.
Another approach is the Mixup technique that enriches the learning data, and has shown success in image classification \cite{verma2019manifold}. Following that, the work GraphMix \cite{verma2021graphmix} proposed an interpolation-based regularization method by parameter sharing of GNNs and point-wise convolutions.

Other methods \textcolor{black}{propose to} consider GNN training from an information and entropy point of view following the success of mutual information in CNNs \cite{hjelm2018learning}. For example, DGI \cite{deepgraphInfomax} learns a global graph vector and considers its correspondence with local patch vectors. However, it does not consider label features as in our work. In the work InfoGraph \cite{sun2019infograph} the authors learn a discriminative network for graph classification tasks, and in \cite{bo2022regularizing_aaai22} a consistency-diversity augmentation is proposed via an entropy perspective for node and graph classification tasks. \textcolor{black}{Another approach to improve the performance of GNNs is to employ multiple self-supervised auxiliary losses \cite{manessi2021graph}, in addition to the task-related supervised loss.}

\subsection{Global information in GNNs}
The use of global information in GNNs was recently proposed in several works. \textcolor{black}{The authors in} \cite{dwivedi2020benchmarkgnns} proposed to utilize the Laplacian eigenvectors as positional encoding, and 
LGL-GNN \cite{li2021lgl} combines graph convolutional learned features with motif-based features to extract both local and global information. Recently, it was also proposed in \cite{huang2022from} to use feature propagation to obtain global features for the node-classification task\textcolor{black}{, and it was also suggested to employ both low and high pass filters that can take advantage of local and global information encoded in the graph structure \cite{wei2022structure}}. \textcolor{black}{Other works like FPGNN \cite{zhang2023fpgnn} and SRGNN \cite{zhang2024learning} study mechanisms to improve the fairness of the graph structure, and \cite{Fu2023HyperbolicGG} study use of hyperbolic graphs representation for imbalanced graphs.} While our work also considers global information, it is different by \textit{what} global information we use. Namely, while the aforementioned methods learn or compute global features that are directly derived from the graph connectivity, here, we aim to learn \textcolor{black}{global }\emph{label} features. \textcolor{black}{Therefore, our method is highly suitable for tasks that involve classification. The key idea is to learn a high dimensional feature per class, that is representative of each class, and to promote \textcolor{black}{the} separation of the classes, as discussed later in Section \ref{sec:method}. As shown by our results in Section \ref{sec:experiments}, and illustrated in Figure \ref{fig:tsne}, the learnt global label features allow to separate the hidden feature space to the desired classes, leading to improved performance in our experiments.}

\subsection{Label information in GNNs} Recently, several works proposed to harness label information to improve the performance of GNNs and to propose novel GNN architectures, \textcolor{black}{ as now discuss}. \textcolor{black}{One example of utilizing label information is} the concept of label propagation \cite{wang2020unifying} suggests that propagating label information according to the graph connectivity is useful for node classification tasks. \textcolor{black}{Another approach, which is presented in } works like \cite{correctAndSmooth} combine label propagation concepts with models like multi-layer perceptron or simple GNNs. In contrast, in this work, we aim to learn label features and utilize them to improve the clustering ability of the network to improve the accuracy of node classification, instead of propagating label information within the graph nodes.



\section{Preliminaries and notations}
\label{sec:notations}
\subsection{Definitions}
We denote an undirected graph by the tuple $\cal G=({\cal V},{\cal E})$  where $\cal V$ is a set of $n$ nodes and $\cal E$ is a set of $m$ edges, 
and by ${\bf f}^{(l)}\in\mathbb{R}^{n \times c}$ the feature tensor of the nodes $\mathcal{V}$ with $c$ \textcolor{black}{dimensions} at the $l$-th layer. We denote the input features by $\textcolor{black}{\bff^{\rm{in}}}$. The adjacency matrix is defined by $\bfA \in \mathbb{R}^{n \times n}$, where $\bfA_{ij} = 1$ if there exists an edge $(i,j) \in {\cal E}$ and 0 otherwise, and the diagonal degree matrix is denoted $\bfD$ where $\bfD_{ii}$ is the degree of the $i$-th node. Furthermore, let us denote the adjacency and degree matrices with added self-edges by $\tilde \bfA$ and $\tilde \bfD$, respectively.

We consider the node classification task with $k$ labels. We denote the ground-truth labels by $\bfy \in \mathbb{R}^{n \times k}$ and the node-classification prediction by applying SoftMax to the output of the network $\bff^{out}$:
\begin{equation}
\label{eq:y_hat}
    \hat{\bfy} = {\rm SoftMax}(\bff^{out}) \in \mathbb{R}^{n \times k}.
\end{equation}

\subsection{GNN Backbones}
\label{subsec:backbones}
In this paper, we utilize three GNN backbones, as described below.

\paragraph{GCN} Using the notations from Section \ref{sec:notations}, the propagation operator from GCN~\cite{kipf2016semi} is obtained by $\tilde{\bfP} = \tilde{\bfD}^{-\frac{1}{2}}\tilde{\bfA}\tilde{\bfD}^{-\frac{1}{2}}$, and its architecture is given by
\begin{equation}
    \label{eq:general_gnn}
    \bff^{(l+1)} = \rm{ReLU}(\tilde{\bfP} \bff^{(l)}\bfK^{(l)}),
\end{equation}
where $\bfK^{(l)}$ is a $1\times1$ convolution matrix.

\paragraph{GAT} The GAT \cite{velickovic2018graph} backbone defines the propagation operator:
\begin{equation}
    \label{eq:attentionCoefficients}
    \alpha_{ij}^{(l)} = \frac{\exp \big({\rm LeakyReLU} \big(\bfa^{(l)^{\top}} [\tilde{\bfK}^{(l)} \bff_i^{(l)} \oplus \tilde{\bfK}^{(l)} \bff_j^{(l)} ]  \big) \big)}{\sum_{p \in \mathcal{N}_i} \exp \big({\rm LeakyReLU} \big(\bfa^{(l)^{\top}} [\tilde{\bfK}^{(l)} \bff_i^{(l)} \oplus \tilde{\bfK}^{(l)} \bff_p^{(l)} ]  \big) \big)},
\end{equation}
where $\bfa^{(l)} \in \mathbb{R}^{2c}$ and $\tilde{\bfK}^{(l)} \in \mathbb{R}^{c \times c}$ are trainable parameters and $\oplus$ denotes channel-wise concatenation and the neighbourhood of the $i$-th node is denoted by $\mathcal{N}_i = \{j | (i,j) \in \mathcal{E}\}$.

By gathering all $\alpha_{ij}^{(l)}$ for every edge $(i,j) \in \mathcal{E}$ into a propagation matrix $\bfS^{(l)} \in \mathbb{R}^{n \times n}$ we obtain the GAT architecture:
\begin{equation}
    \bff^{(l+1)} = \rm{ReLU}({\bfS^{(l)}} \bff^{(l)}\bfK^{(l)}).
\end{equation}

\paragraph{GCNII} \textcolor{black}{In GCNII \cite{chen20simple}, the feature propagation is defined as}:
\begin{equation}
\bfS^{(l)} (\bff^{(l)}, \bff^{(0)})=   (1-\alpha^{(l)}) \tilde{\bfP}\bff^{(l)} + \alpha^{(l)}\bff^{(0)},
\end{equation}
where $\alpha^{(l)} \in [0, 1]$ is a hyper-parameter and $\bff^{(0)}$ are the features of the first (embedding) layer. Then, a GCNII layer is given by:

\begin{equation}    
    \bff^{(l+1)} = \sigma(\beta^{(l)}\bfS^{(l)} (\bff^{(l)}, \bff^{(0)})\bfK^{(l)}  + (1-\beta^{(l)}) \bfS^{(l)} (\bff^{(l)},  \bff^{(0)})),
\end{equation}    
where $\beta^{(l)} \in [0,1]$ is a hyper-parameter. Later, in our experiments, we will use the same values as reported in \cite{chen20simple} for $\alpha^{(l)}$ and $\beta^{(l)}$, \textcolor{black}{and $\sigma$ is an activation function.}

\section{Method}
\label{sec:method}

\begin{figure*}[t]
    \centering    \includegraphics[width=0.65\linewidth]{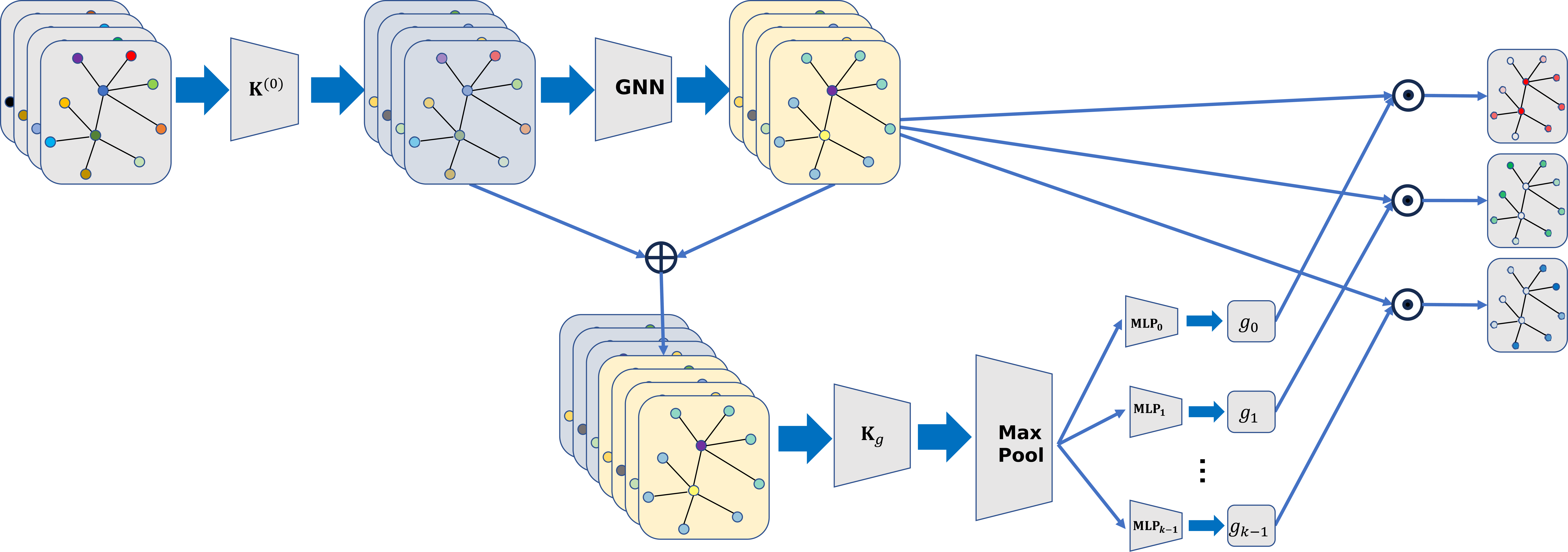}
	\captionof{figure}{\textcolor{black}{Our GLGNN architecture. We learn node and label features, and fuse them to obtain a response per node for each label. $\oplus$ denotes feature-wise concatenation. $\odot$ denotes matrix-vector dot product. $\rm{MLP}_q$ implements Equation \eqref{eq:SE} for $0 \leq q \leq k-1$.}}
	\label{fig:architecture}
\end{figure*}

\subsection{Local node features}
\textcolor{black}{In general,} the local information is obtained by learning node features $\bff \in \mathbb{R}^{n \times d}$ using some backbone GNN. In our experiments, we use three different backbones to demonstrate the benefit of incorporating global information. Specifically, we employ GCN \cite{kipf2016semi}, GAT \cite{velickovic2018graph} and GCNII \cite{chen20simple} as described in Section \ref{subsec:backbones}. Note that our GLGNN approach does not assume a specific GNN backbone and thus can possibly be applied to other GNNs.

\subsection{Global label features}
Our global information mechanism learns label features $\bfg \in \mathbb{R}^{k \times d}$. \textcolor{black}{The goal is to learn label features that characterize the different classes in the data, and will be used in conjunction with the learnt node features to obtain classification prediction maps.} Specifically, to obtain the global features, we consider the concatenation of initial node-embedding $\bff^{(0)} = \textcolor{black}{\bff^{\rm{in}}} \bfK^{(0)}$, and the last $\rm{GNN}$ layer node features $\bff^{(\rm{L})}$ denoted by $\left[ \bff^{(0)} \oplus \bff^{({\rm{L}})} \right]$. We then perform a single $1\times1$ convolution denoted by $\bfK_{\rm{g}}$, followed by a ReLU activation \cite{fukushima1975cognitron}, and feed it to a global MaxPool readout function to obtain a single vector $\bfs \in \mathbb{R}^d$. Formally:
\begin{equation}
    \label{eq:globalReadout}
    \bfs  = \rm{MaxPool}\left(\rm{ReLU}\left(\bfK_{g}\left[ \bff^{(0)} \oplus \bff^{(L)} \right]\right)\right).
\end{equation}
Using the global vector $\bfs$, we utilize $k$ (the number of labels) multi-layer perceptrons (MLPs), \cite{rosenblatt1962principles}, as follows: 
\begin{equation}
    \label{eq:SE}
    {\bf{g}}_{\textcolor{black}{q}} = {\bfK}_{s}^{\textcolor{black}{(q)}}\left(\rm{ReLU}\left(\bfK_{e}^{\textcolor{black}{(q)}}\bfs \right)\right),
\end{equation}
where $\bfK_{\rm{e}}^{\textcolor{black}{(q)}}, \bfK_{\rm{s}}^{\textcolor{black}{(q)}}$ are an expanding (from $d$ to $e\times d$) and shrinking (from $e \times d$ to $d$) $1\times 1$ convolutions, \textcolor{black}{$q$ denotes the $q$-th class}, and the expansion rate $e$ is a hyper-parameter which is set $e=12$ in our experiments. \textcolor{black}{The channel expansion multiplies the feature dimensionality and has been shown to be effective in CNNs \cite{sandler2018mobilenetv2}.} Note that Equation \eqref{eq:SE} can be implemented using a grouped convolution to obtain $\bfg = [\bfg_{\rm{0}}, \ldots , \bfg_{\rm{k-1}}]$ in parallel. \textcolor{black}{We illustrate the proposed learning mechanism in Figure \ref{fig:architecture}.} 

\subsection{Node-classification map using global label features} To obtain a node-classification prediction map, we consider the matrix-vector product of the final $\rm{GNN}$ output $\bff^{(L)} \in \mathbb{R}^{n \times d}$ with each of the label features $\bfg_{\textcolor{black}{q}} \in \mathbb{R}^{d}$ in \eqref{eq:SE}. More formally, for each label, we obtain the following node-label correspondence vector:
\begin{equation}
    \label{eq:nodeLabelDotProduct}
    \bfz_{\textcolor{black}{q}} = \bff^{(L)}\cdot \bfg_{\textcolor{black}{q}} \in \mathbb{R}^{n} \textcolor{black}{\ , \ q=0, \ldots,k-1}
\end{equation}
By concatenating the $k$ correspondence vectors and applying the SoftMax \cite{bridle1989training} function, we obtain a node-classification map,  the final output of our GLGNN is given by
\begin{equation}
    \label{eq:predictionMap}
    \hat{\bfy} = \rm{SoftMax}\left(\bfz_{0} \oplus \ldots \oplus \bfz_{k-1} \right) \in \mathbb{R}^{n \times k}.
\end{equation}

\subsection{Objective functions} To train  GLGNN  minimize the objective function:
\begin{equation}
    \label{eq:totalObjective}
    \mathcal{L} = \mathcal{L}_{CE} + \gamma \mathcal{L}_{\rm{GL}},
\end{equation}
where $\mathcal{L}_{CE}$ denotes the cross-entropy loss between ground-truth $\bfy$ and predicted node labels $\hat{\bfy}$ from Equation \eqref{eq:predictionMap}, formally defined as follows: 
\begin{equation}
    \label{eq:crossEntropy}
    \mathcal{L}_{CE} = - \frac{1}{\mid \mathcal{V}^{lab}\mid}\sum_{i \in \mathcal{V}^{lab}}\sum_{s=0}^{k-1}  \bfy_{i,s} \log(\hat{\bfy}_{i,s}),
\end{equation} 
where the set of labelled nodes is denoted by $\mathcal{V}^{lab}$.
$\gamma$ is a positive hyper-parameter, and $\mathcal{L}_{\rm{GL}}$ denotes a global-local loss that considers the relationship between the label and node features by demanding the similarity of nodes that belong to a respective label while requiring the dis-similarity of node features that do not belong to that label and its features, as follows
\begin{equation}
    \label{eq:globalLocalLoss}
    \mathcal{L}_{\rm{GL}} =  \sum_{{\textcolor{black}{q}}=0}^{k-1}  \left( \sum_{\bfy_{i}={\textcolor{black}{q}}} \|\bfg_{{\textcolor{black}{q}}} - \bff^{(L)}_{i}\|_{2}^{2} - \sum_{\bfy_{i} \neq {\textcolor{black}{q}}} \rm{min}\left(\| \bfg_{{\textcolor{black}{q}}} - \bff^{(L)}_{i} \|_{2}^{2}, r \right) \right),
\end{equation}
where $\rm{min(\cdot, \cdot)}$ is a clamping function that returns the minimal values of its arguments, and $\rm{r}$ is a positive hyper-parameter. \textcolor{black}{The clamping function $\rm{min(\cdot, \cdot)}$ is utilized to set a maximal value on the out-of-class distances, which otherwise can be infinity. This is a common approach when utilizing contrastive losses \cite{khosla2020supervised}}. In our experiments, we choose $\rm{r}=10$ \textcolor{black}{as it provided the overall highest accuracy across the datasets in this paper}. We also perform an ablation study of $\rm{r}$ in Section \ref{sec:ablation}.

\subsection{Computational complexity and costs}
We now analyse the complexity of the  considered backbones of GCN
 \cite{kipf2016semi}, GAT \cite{velickovic2018graph}, and GCNII \cite{chen20simple} and then analyse the complexity of our added global information module. We provide the number of parameters, FLOPs, and runtimes in Table \ref{tab:runtimes}. We see that our GLGNNs do not require a significant additional computational cost, at the return of favourable performance.

\paragraph{Backbone}
In all backbones, an embedding layer that consumes the input features is used, which requires $n \times c_{in} \times c$ FLOPs.
The GCN and GCNII backbones with $\rm{L}$ layers require $n \times c \times c \times \rm{L}$ FLOPs. The GAT \cite{velickovic2018graph} backbone requires $(n \times c \times c + n \times \tilde{d} \times c) \times \rm{L}$ FLOPs. The added cost of GAT follows from the pair-wise score computations, where $\tilde{d}$ is the average degree of the graph.

\paragraph{Global information module} The global label information module of our GLGNN is described by the computations in Equation \eqref{eq:globalReadout} and \eqref{eq:SE} requires $n \times 2c \times c$ and $2 \times e \times c \times k$ FLOPs, respectively. Most importantly, the label features are computed past the global readout in Equation \eqref{eq:globalReadout} and therefore are not dependent on the number of nodes $n$. Therefore it maintains a relatively low computational overhead as depicted \textcolor{black}{in} Table \ref{tab:runtimes}.

\begin{table}[]
  \caption{Million floating-point operations (FLOPs [M]) and GPU runtimes in milliseconds (ms) on \textcolor{black}{Cora \cite{mccallum2000automating}} \cite{} with 64 \textcolor{black}{dimensions}, expansion of $e=12$ and 2-layer backbones.}
  \label{tab:runtimes}
  \begin{center}
  \resizebox{\linewidth}{!}{
  \begin{tabular}{lcccccc}
  \toprule
   Metric  & GCN & GAT   & GLGCN (Ours) & GLGAT (Ours)  \\
    \midrule
    FLOPs [M] & 271.7 & 273.4 & 294.3 &  295.9   \\
    Training [ms] & 7.80 & 14.48  & 8.31 & 15.02  \\
    Inference [ms]  & 2.75 & 3.34   &  3.23 & 3.96 \\
    Test accuracy ($\%$) &  81.1 & 83.1  & 84.2 & 84.5 \\
    \bottomrule
  \end{tabular}}
  \end{center}
\end{table}

\section{Implementation details}
\label{appendix:architectures}
In this section, we provide details about the architecture used in our experiments in Section 
\ref{sec:experiments}.
Our network architecture consists of an opening (embedding) layer ($1
\times 1$ convolution), a sequence of GNN backbones layers, and a series of $1\times1$ convolutions to learn the global labels features.
In our experiments, we employ GLGCN, GLGAT, and GLGCNII, which differ by the GNN backbone, as discussed in Section 
\ref{sec:method}. We specify the node feature backbone architecture in Table \ref{table:nodeClassificationArch}, and the global label feature extraction architecture in Table \ref{table:labelFeaturesArch}. In what follows, we denote by $c_{in}$ and $k$ the input and output \textcolor{black}{dimensions}, respectively, and $c$ denotes the number of features in hidden layers. We initialize the embedding and label features-related layers with the Glorot \cite{glorot2010understanding} 
initialization, and $\bfK^{(l)}$ from Equation \eqref{eq:general_gnn} is initialized with an identity matrix of shape $c \times c$. We denote the number of GNN layers by $\rm{L}$, and the dropout probability by $\rm{p}$. In all of the experiments with GLGCN and GLGAT we use 2 layers, as those backbones will over-smooth \cite{measuringoversmoothing}. With GLGCNII we use the same number of layers as reported in \cite{chen20simple}.


\begin{table}[]
    \centering
    \small
  \caption{The architecture used for node features extraction.}
  \label{table:nodeClassificationArch}

  \begin{tabular}{lcc}
  \toprule
    Input size & Layer  &  Output size \\
    \midrule
    $n \times c_{in}$ & Dropout(p) & $n \times c_{in}$ \\
    $n \times c_{in}$ & $1\times1$ Convolution & $n \times c$ \\
    $n \times c$ & ReLU & $n \times c$ \\    
    $n \times c$ & ${\rm{L}} \times $ GNN backbone & $n \times c$ \\
    \bottomrule
  \end{tabular}
\end{table}

\begin{table}[t]
    \centering
    \small
  \caption{The architecture used for label features extraction. The input of this architecture is the output of Table \ref{table:nodeClassificationArch}}
    \label{table:labelFeaturesArch}
    \begin{tabular}{lcc}
  \toprule
    Input size & Layer  &  Output size \\
    \midrule
    $n \times c$ & MaxPool & $1 \times c$ \\
    $1 \times c$ & $k \times 1\times1$ Convolutions & $k \times e \cdot c$ \\
    $k \times e \cdot c$ & ReLU & $k \times e \cdot c$ \\    
    $k \times e\cdot c$ & $k \times 1\times1$ Convolutions & $k \times  c$ \\    \bottomrule
  \end{tabular}
\end{table}

\begin{table}[]
    \centering
    \small
  \caption{Grid search ranges. LR and WD denote the learning rate and weight decay of embedding and label feature layers. LR\textsubscript{GNN} and WD\textsubscript{GNN} denote the learning rate and weight decay of GNN layers.}
  \label{table:hyperparams}
   
  \begin{tabular}{lc}
  \toprule
    Hyper-parameter & Values range\\
    \midrule
    LR / LR\textsubscript{GNN}  & [1e-1, 1e-2, 1e-3, 1e-4] \\
    WD / WD\textsubscript{GNN} & [1e-3, 1e-4, 1e-5, 0]\\
    $\gamma$ & [1e+2, 1e+1,1, 1e-1,1e-2]\\
    $\rm{p}$ & [0.5,0.6,0.7] \\
   \bottomrule
  \end{tabular}
\end{table}

\begin{table}[t]
    \centering
    \small
      \caption{Datasets statistics.}
  \label{table:datasets}

  \begin{tabular}{lcccc}
  \toprule
    Dataset & Classes & Nodes & Edges & Features  \\
    \midrule
    Cora \cite{mccallum2000automating} & 7 & 2,708 & 5,429 & 1,433 \\
    Citeseer \cite{sen2008collective} & 6 & 3,327  & 4,732 & 3,703 \\
    Pubmed \cite{namata2012query} & 3 & 19,717 & 44,338 & 500 \\
    Cornell \cite{Pei2020Geom-GCN:} & 5 & 183 & 295 & 1,703 \\
    Texas \cite{Pei2020Geom-GCN:} & 5 & 183 & 309 & 1,703 \\
    Wisconsin \cite{Pei2020Geom-GCN:} & 5 & 251 & 499 & 1,703 \\
    Chameleon \cite{musae} & 5 & 2,277 &  36,101 & 2,325\\
    Actor \cite{Pei2020Geom-GCN:} & 5 & 7,600 & 33,544 & 932  \\
    Squirrel \cite{musae} & 5 & 5,201 & 198,493 &  2,089 \\
    \bottomrule
  \end{tabular}
\end{table}

\section{Experiments}
\label{sec:experiments}

We now demonstrate GLGNN on semi and fully  supervised node classification. As GNN backbones, we consider GCN \cite{kipf2016semi}, GAT \cite{velickovic2018graph}, and GCNII \cite{chen20simple}. In all experiments, we use the Adam~\cite{kingma2014adam} optimizer, and perform a grid search to choose the hyper-parameters. The range of values that were tested in the grid search is reported in Table \ref{table:hyperparams}. Our code is implemented using PyTorch~\cite{pytorch}, trained on an Nvidia\textsuperscript{\textregistered} Titan RTX GPU \cite{burgess2020rtx}.

We show that for all the considered node classification datasets, whose statistics are provided in Table \ref{table:datasets}, our GLGNNs offer a consistent improvement over the baseline methods, and besides the obtained accuracy we report the relative accuracy improvement compared to the baseline methods, and competitive with recent state-of-the-art methods.

\subsection{Semi-Supervised Node-Classification}
\label{sec:semisupervised}
\begin{figure}[t]
    \centering    \includegraphics[width=0.50\linewidth]{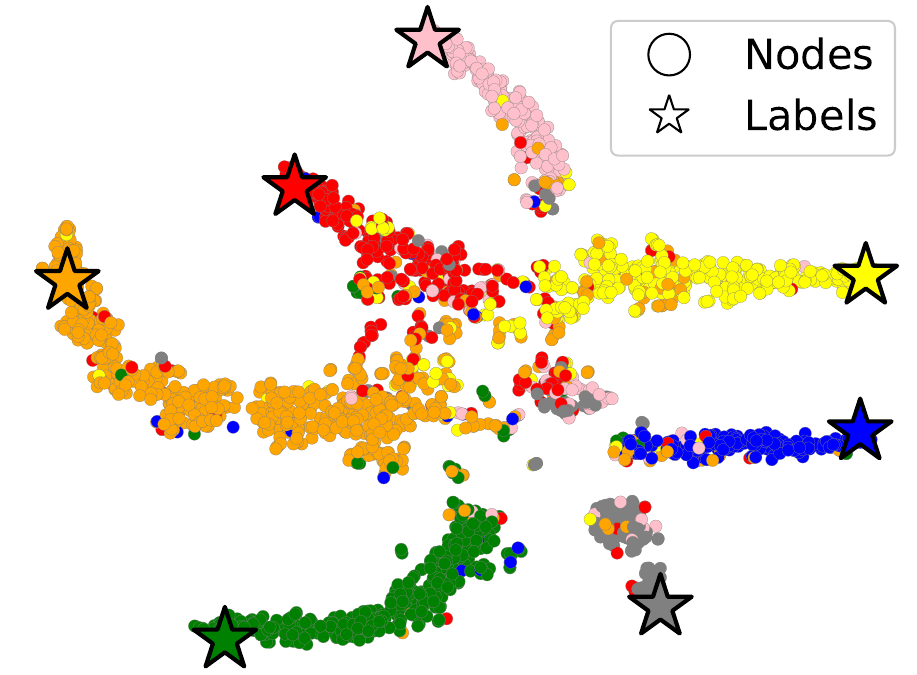}
	\captionof{figure}{tSNE embedding of learnt label- and node-features of Cora. The similarity of the label features and the corresponding node features shows the clustering effect of incorporating global information.}
	\label{fig:tsne}
\end{figure}

We consider Cora \cite{mccallum2000automating}, Citeseer  \cite{sen2008collective} and Pubmed \cite{namata2012query} datasets and their standard, public training/validation/testing split as in~\cite{yang2016revisiting}, with 20 nodes per class for training.
We follow the training and evaluation scheme of~\cite{chen20simple} and consider various GNN models like GCN, GAT, superGAT~\cite{kim2020findSuperGAT}, APPNP~\cite{klicpera2018combining}, JKNet~\cite{jknet}, GCNII \cite{chen20simple}, GRAND~\cite{chamberlain2021grand}, PDE-GCN~\cite{eliasof2021pde}, pathGCN \cite{eliasof2022pathgcn}, EGNN\cite{zhou2021dirichlet} and superGAT~\cite{kim2020findSuperGAT}. We also consider other improved training techniques P-reg~\cite{yang2021rethinking}, GraphMix~\cite{verma2021graphmix}, and NASA~\cite{bo2022regularizing_aaai22}. We summarize the results in Table~\ref{table:semisupervised_summary} and illustrate the learnt labels and nodes features in Figure \ref{fig:tsne}, revealing the clustering effect of learning label nodes. Our results show that by incorporating learnable global label information, our GLGNN variants consistently improve the baseline architectures of GCN, GAT, and GCNII, and also offer competitive accuracy that is in line with other state-of-the-art methods. For example, our GLGCNII obtains 83.7\% accuracy on the Pubmed dataset, an improvement of 3.4\% over the baseline GCNII and 1.0\% advantage over the second highest method, pathGCN, with 82.7\% accuracy. 

\begin{table}[]
    \centering
    \small
  \caption{Semi-supervised node-classification accuracy (\%). \textcolor{black}{In parentheses we present the accuracy gain compared to the backbones.}}
  \label{table:semisupervised_summary}
  \begin{tabular}{lccc}
  \toprule
    Method & Cora & Citeseer & Pubmed \\ 
    \midrule
    GCN \cite{kipf2016semi} & 81.1 & 70.8 & 79.0  \\
    GAT \cite{velickovic2018graph} & 83.1 & 70.8 & 78.5  \\
    GCNII \cite{chen20simple} & 85.5 & 73.4 & 80.3  \\
    \midrule
    APPNP \cite{klicpera2018combining} & 83.3 & 71.8 & 80.1 \\
    JKNET \cite{jknet} & 81.1 & 69.8 & 78.1 \\
    GRAND \cite{chamberlain2021grand} & 84.7 & 73.6 & 81.0  \\
    PDE-GCN \cite{eliasof2021pde} & 84.3 & 75.6 & 80.6  \\
    pathGCN \cite{eliasof2022pathgcn} & 85.8 & 75.8 & 82.7 \\
    EGNN \cite{zhou2021dirichlet} & 85.7 & -- & 80.1 \\
    superGAT \cite{kim2020findSuperGAT} & 84.3 & 72.6 & 81.7 \\
    GraphMix \cite{verma2021graphmix} & 84.0 & 74.7 & 81.1 \\ 
    P-reg \cite{yang2021rethinking} & 83.9 & 74.8 & 80.1 \\
    NASA \cite{bo2022regularizing_aaai22} & 85.1 & 75.5 & 80.2  \\
    \midrule
    GLGCN (ours) & 84.2${\scriptscriptstyle (+\textcolor{black}{3.1\%})}$ & 73.3 ${\scriptscriptstyle (+\textcolor{black}{2.5\%})}$ & 81.5 ${\scriptscriptstyle (+\textcolor{black}{2.5\%})}$  \\
    GLGAT (ours) & 84.5${\scriptscriptstyle (+\textcolor{black}{1.6\%})}$ & 72.6${\scriptscriptstyle (+\textcolor{black}{1.8\%})}$ & 81.2${\scriptscriptstyle (+\textcolor{black}{2.7\%})}$  \\
        GLGCNII (ours) & 85.7${\scriptscriptstyle (+0.2\%)}$ & 75.4${\scriptscriptstyle (+2.0\%)}$ & 83.7${\scriptscriptstyle (+3.4\%)}$  \\
    \bottomrule
    \end{tabular}
\end{table}

\subsection{Fully-Supervised Node-Classification}
\label{sec:fullysupervised}

\begin{table}[t]
    \centering
    \small
  \caption{Fully-supervised node-classification accuracy ($ \%$) on \emph{homophilic} datasets. \textcolor{black}{In parentheses we present the accuracy gain compared to the backbones.} \textcolor{black}{GCNII* is a second variant of GCNII proposed in the original paper \cite{chen20simple}.}}
  \label{table:homophilic_fully}
  \begin{tabular}{lccc}
    \toprule
    Method & Cora & Citeseer & Pubmed \\
    Homophily & 0.81 & 0.80 & 0.74 \\
        \midrule
    GCN  & 85.77 & 73.68 & 88.13  \\
    GAT & 86.37 & 74.32 & 87.62 \\
    GCNII & 88.49  & 77.08 & 89.57  \\
    \midrule
    Geom-GCN & 85.27 & 77.99 & 90.05\\
    APPNP &  87.87 & 76.53 & 89.40 \\
    JKNet & 85.25  & 75.85  & 88.94 \\
    JKNet (Drop) & 87.46  & 75.96 & 89.45 \\
    WRGAT & 88.20 & 76.81 & 88.52 \\
    Incep (Drop) & 86.86 & 76.83 & 89.18 \\
    GCNII*  & 88.01 & 77.13 & 90.30 \\
    GGCN  & 87.95  & 77.14   & 89.15 \\
    H2GCN  & 87.87  & 77.11   & 89.49 \\
    C\&S & 89.05 & 77.06 & 89.74 \\
    GPRGNN & 87.95 & 77.13 & 87.54 \\
    PowerEmbed & 85.03 & 73.27 & --  \\
    \midrule
    GLGCN (ours) & 88.47${\scriptscriptstyle (+\textcolor{black}{2.7\%})}$ & 77.72 ${\scriptscriptstyle (+\textcolor{black}{4.0\%})}$ & 88.61 ${\scriptscriptstyle (+\textcolor{black}{0.4\%})}$ \\
    GLGAT (ours) & 88.65${\scriptscriptstyle (+\textcolor{black}{2.2\%})}$ & 77.37 ${\scriptscriptstyle (+\textcolor{black}{3.0\%})}$ & 88.74 ${\scriptscriptstyle (+\textcolor{black}{1.1\%})}$  \\
    GLGCNII (ours) & 89.43$ \scriptscriptstyle{(+0.9\%)}$ & 79.16$ \scriptscriptstyle{(+2.0\%)}$ & 90.83 $\scriptscriptstyle(+1.2\%)$  \\
    \bottomrule
  \end{tabular}
\end{table}

\begin{table*}[t]
\centering
\footnotesize
  \caption{Fully-supervised node-classification accuracy ($ \%$) on \emph{heterophilic} datasets. \textcolor{black}{In parentheses we present the accuracy gain compared to the backbones.}}
  \label{table:heterophilic_fully2}
  \begin{tabular}{lcccccc}
    \toprule
    Method & Cornell & Wisconsin & Texas & Cham.  & Squirrel & Actor\\
    Homophily & 0.30 & 0.21 & 0.11 & 0.23 &  0.22 & 0.22 \\
        \midrule
    GCN &  52.70 & 52.16 & 48.92  & 28.18 & 23.96 & 27.32 \\
    GAT  & 54.32 & 58.38 & 49.41 & 42.93 & 30.03 & 27.44
    \\
        GCNII  & 74.86 & 69.46 & 74.12   & 60.61 & 38.47 & 32.87 \\
    \midrule
    Geom-GCN  & 60.81  & 67.57 & 64.12  & 60.90 & 38.32 & 31.63\\
    APPNP  &   54.30 & 73.51  & 65.41   & 51.91 & 34.77 & 38.86  \\
    PairNorm   &  58.92 & 60.27 & 48.43 &  62.74 & 50.44 & 27.40  \\
    GCNII* &   76.49 & 77.84  & 81.57  & 62.48 & 39.92 & 33.61   \\
    GRAND  &   82.16 & 75.68 & 79.41 & 54.67 & 40.05 & 35.62 \\ 
    WRGAT  &  81.62 & 83.62 & 86.98 & 65.24 & 48.85 & 36.53 \\
    GGCN   & 85.68  & 84.86  &  86.86   & 71.14 & 55.17 & 37.81  \\ 
    H2GCN   & 82.70  & 84.86  &  87.65  & 60.11 & 36.48 & 35.70 \\
    FAGCN & 79.19 & 82.43 & 82.94 & 55.22 & 42.59 & 34.87 \\
    GPRGNN & 80.27 & 78.38 & 82.94 & 46.58 & 31.61 & 34.63  \\
    PowerEmbed & 78.30 & 78.43 & 79.19 & 64.98 & 53.53 & -- \\
    \midrule
    GLGCN (ours) & 74.86 ${\scriptscriptstyle (+\textcolor{black}{22.1\%})}$ &  70.27 ${\scriptscriptstyle (+\textcolor{black}{18.1\%})}$  & 65.29  ${\scriptscriptstyle (+\textcolor{black}{16.3\%})}$ & 65.17$ \scriptscriptstyle{(+36.9\%)}$ & 51.64  $ \scriptscriptstyle{(+27.6\%)}$& 32.45 $ \scriptscriptstyle{(+\textcolor{black}{5.1\%})}$  \\
    GLGAT (ours)  & 75.67 ${\scriptscriptstyle (+\textcolor{black}{21.4\%})}$  & 70.01 ${\scriptscriptstyle (+\textcolor{black}{11.6\%})}$  & 65.88 ${\scriptscriptstyle (+\textcolor{black}{16.4\%})}$  & 65.24 $ \scriptscriptstyle{(+\textcolor{black}{22.3\%})}$ & 52.59 $ \scriptscriptstyle{(+22.5\%)}$& 31.84 $ \scriptscriptstyle{(+4.4\%)}$  \\
    GLGCNII (ours)  & 89.19 $\scriptscriptstyle{(+14.3\%)}$ & 89.72 $\scriptscriptstyle{(+20.2\%)}$ & 88.62 $\scriptscriptstyle{(+14.5\%)}$ & 66.62 $ \scriptscriptstyle{(+6.0\%)}$ & 52.52 $ \scriptscriptstyle{(+14.0\%)}$ & 34.13 $ \scriptscriptstyle{(+1.2\%)}$  \\
    \bottomrule
  \end{tabular}
\end{table*}

\begin{table}[]
    \centering
    \footnotesize
  \caption{\textcolor{black}{Graph classification accuracy (\%).}}
  \label{table:graphClassification}
  \begin{tabular}{lccc}
  \toprule
    Method & MUTAG & PROTEINS & PTC \\ 
    \midrule
    GCN & 85.6 $\pm$ 5.8 & 76.0 $\pm$ 3.2 & 64.2 $\pm$ 4.3  \\
    GIN & 89.0 $\pm$ 6.0 & 75.9 $\pm$ 3.8 & 63.7 $\pm$ 8.2  \\
    \midrule
    GLGCN (ours) & 88.3 $\pm$ 5.1 & 76.8 $\pm$ 3.3 & 65.8 $\pm$ 4.0  \\ 
    GLGIN (ours) & 90.8 $\pm$ 6.1 & 76.4 $\pm$ 3.5 & 66.0 $\pm$ 2.8  \\
    \bottomrule
    \end{tabular}
\end{table}

\begin{table}[t]
\centering
\footnotesize
  \caption{Accuracy (\%) vs. use of the initial node features $\bff^{(0)}$.}
  \label{table:initresidual}
  \begin{tabular}{lccc}
    \toprule
    Init. & Cora & Citeseer & Pubmed \\
     Feat. & & &  \\
        \midrule
    \checkmark  & 84.2 & 73.3 & 81.5  \\
    \midrule
    X & 83.8 & 72.3  &  81.0 \\
    \bottomrule
    \end{tabular}
\end{table}

To further validate the efficacy of our method, we experiment with the fully supervised node classification task on 9 datasets, namely, Cora, Citeseer, Pubmed, Chameleon, Squirrel, Actor (Film), Cornell, Texas and Wisconsin using the 10 random splits from~\cite{Pei2020Geom-GCN:} with train/validation/test split of $48 \%, 32\%, 20\%$ respectively, and report their average accuracy. In all experiments, we use 64 \textcolor{black}{dimensions} and perform a grid search to determine the hyper-parameters. We compare our accuracy with methods like GCN, GAT, Geom-GCN~\cite{Pei2020Geom-GCN:}, APPNP, JKNet~\cite{jknet}, WRGAT~\cite{Suresh2021BreakingTL}, GCNII \cite{chen20simple}, DropEdge \cite{Rong2020DropEdge:}, H2GCN~\cite{zhu2020beyondhomophily_h2gcn}, GGCN~\cite{yan2021two}, C\&S \cite{correctAndSmooth}, FAGCN
\cite{bo2021beyond}, GPRGNN \cite{chien2021adaptive}, and PowerEmbed \cite{huang2022from}. We report our accuracy on homophilic and heterophilic datasets in Tabs. \ref{table:homophilic_fully} and \ref{table:heterophilic_fully2}, respectively. We also report the homophily score of each dataset based on \cite{Pei2020Geom-GCN:}. We see an improvement across all types of datasets compared to the baseline methods, and competitive results with recent state-of-the-art methods on homophilic datasets.

\subsection{\textcolor{black}{Graph classification}}
\label{sec:graphClassification}
\textcolor{black}{The main focus of our GLGNN approach is to improve the performance on node classification tasks, by combining both learnt node and label features. However, GLGNN can also be utilized on other classification tasks such as graph classification. To obtain a graph-level prediction, after obtaining the node and class representations illustrated in Figure \ref{fig:architecture}, we perform an add-pool operation, followed by a standard classifier, as in GIN \cite{xu2018how}. We now report and compare our GLGNN with GCN and GIN as backbones and baseline networks on the MUTAG, PROTEINS, and PTC datasets from \cite{Morris2020TUDatasets}. We \textcolor{black}{follow} the same training and testing procedures and protocols from \cite{xu2018how}. The results are provided in Table \ref{table:graphClassification}, where we see that GLGCN and GLGIN offer improved accuracy compared to the baseline models.}

\subsection{Ablation study}
\label{sec:ablation}

\begin{figure}[t]

    \centering
\begin{subfigure}[t]{.48\linewidth}
    \centering
    \begin{tikzpicture}
      \begin{axis}[
          width=1.0\linewidth, 
          height=0.8\linewidth,
          grid=major,
          grid style={dashed,gray!30},
          xlabel=Channel expansion $\rm{e}$,
          ylabel=Accuracy (\%),
          ylabel near ticks,
        legend style={at={(1.45,1.5)},anchor=north,scale=1.0, cells={anchor=west}, fill=none},
          legend columns=3,
          xtick={1,2,4,8,12,16,32},
          xticklabels = {1,2,4,8,12,16,32},
          yticklabel style={
            /pgf/number format/fixed,
            /pgf/number format/precision=3
          },
          ticklabel style = {font=\tiny},
          scaled y ticks=false,
          every axis plot post/.style={thick},
        ]
        \addplot
        table[x=expansion
,y=cora,col sep=comma] {expansion_value.csv};
        \addplot
        table[x=expansion
,y=citeseer,col sep=comma] {expansion_value.csv};
        \addplot
        table[x=expansion
,y=pubmed,col sep=comma] {expansion_value.csv};
        \legend{Cora, Citeseer, Pubmed}
        \end{axis}
    \end{tikzpicture}
                \caption{}
        \label{fig:expansion}
    \end{subfigure}
    \hfill
    \begin{subfigure}[t]{.48\linewidth}
    \centering
    \begin{tikzpicture}
      \begin{axis}[
          width=1.0\linewidth, 
          height=0.8\linewidth,
          grid=major,
          grid style={dashed,gray!30},
          xlabel=Cut-off value $\rm{r}$,
          ylabel=Accuracy (\%),
          ylabel near ticks,
          legend style={at={(0.8,0.65)},anchor=north,scale=1.0, draw=none, cells={anchor=west}, font=\tiny, fill=none},
          legend columns=1,
          xtick={0.01, 0.1, 1, 10, 100,1000,10000},
          xticklabels = {$10^{-2}$, $10^{-1}$, $10^{0}$, $10^{1}$, $10^{2}$, $10^{3}$, $10^{4}$},
          yticklabel style={
            /pgf/number format/fixed,
            /pgf/number format/precision=3
          },
          ticklabel style = {font=\tiny},
          xmode=log,
          scaled y ticks=false,
          every axis plot post/.style={thick},
        ]
        \addplot
        table[x=cutoff
,y=cora,col sep=comma] {cutoff_value.csv};
        \addplot
        table[x=cutoff
,y=citeseer,col sep=comma] {cutoff_value.csv};
        \addplot
        table[x=cutoff
,y=pubmed,col sep=comma] {cutoff_value.csv};
        \end{axis}
    \end{tikzpicture}
                    \caption{}
        \label{fig:cuttoff}
    \end{subfigure}%
    \\

        \begin{subfigure}[t]{.48\linewidth}
    \centering
    \begin{tikzpicture}
      \begin{axis}[
          width=1.0\linewidth, 
          height=0.8\linewidth,
          grid=major,
          grid style={dashed,gray!30},
          xlabel=Dropout value p (\%),
          ylabel=Accuracy (\%),
          ylabel near ticks,
          legend style={at={(0.8,0.65)},anchor=north,scale=1.0, draw=none, cells={anchor=west}, font=\tiny, fill=none},
          legend columns=1,
          xtick={0.0, 0.1, 0.2, 0.3, 0.4,0.5,0.6, 0.7},
          xticklabels = {0.0, 0.1, 0.2, 0.3, 0.4,0.5,0.6, 0.7},
          yticklabel style={
            /pgf/number format/fixed,
            /pgf/number format/precision=3
          },
          ticklabel style = {font=\tiny},
          scaled y ticks=false,
          every axis plot post/.style={thick},
        ]
        \addplot
        table[x=cutoff
,y=cora,col sep=comma] {dropout_value.csv};
        \addplot
        table[x=cutoff
,y=citeseer,col sep=comma] {dropout_value.csv};
        \addplot
        table[x=cutoff
,y=pubmed,col sep=comma] {dropout_value.csv};
        \end{axis}
    \end{tikzpicture}
                    \caption{}
        \label{fig:dropout}
    \end{subfigure}%
    \hfill
        \begin{subfigure}[t]{.48\linewidth}
    \centering
    \begin{tikzpicture}
      \begin{axis}[
          width=1.0\linewidth, 
          height=0.8\linewidth,
          grid=major,
          grid style={dashed,gray!30},
          xlabel= $\gamma$ value,
          ylabel=Accuracy (\%),
          ylabel near ticks,
          legend style={at={(0.8,0.65)},anchor=north,scale=1.0, draw=none, cells={anchor=west}, font=\tiny, fill=none},
          legend columns=1,
          xtick={0.01, 0.1, 1, 10, 100},
          xticklabels = {$10^{-2}$, $10^{-1}$, $10^{0}$, $10^{1}$, $10^{2}$,},
          yticklabel style={
            /pgf/number format/fixed,
            /pgf/number format/precision=3
          },
          ticklabel style = {font=\tiny},
          xmode=log,
          scaled y ticks=false,
          every axis plot post/.style={thick},
        ]
        \addplot
        table[x=cutoff
,y=cora,col sep=comma] {gamma_value.csv};
        \addplot
        table[x=cutoff
,y=citeseer,col sep=comma] {gamma_value.csv};
        \addplot
        table[x=cutoff
,y=pubmed,col sep=comma] {gamma_value.csv};
        \end{axis}
    \end{tikzpicture}
                    \caption{}
        \label{fig:gammaValue}
    \end{subfigure}%
\caption{Ablation study accuracy (\%) on semi-supervised node classification \textcolor{black}{using our GLGCN} vs. (a) the channel expansion rate $\rm{e}$ from Equation \eqref{eq:SE}.  (b) the cut-off value $\rm{r}$ from Equation \eqref{eq:globalLocalLoss}. {(c) The dropout value $p$. (d) The value of $\gamma$ from Equation \eqref{eq:totalObjective}.}}
\label{fig:ablationFigs}
\end{figure}
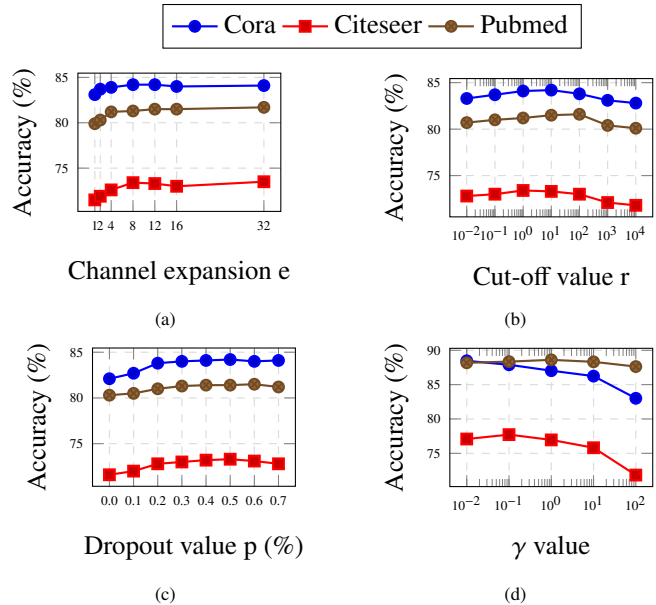

We perform three additional experiments to better understand the different components and behaviour of our GLGNNs.
\paragraph{The use of the initial node features term in Equation \eqref{eq:globalReadout}} In Table \ref{table:initresidual} we report the obtained accuracy and Cora, Citeseer, and Pubmed in the semi-supervised settings as in Section \ref{sec:semisupervised}, with and without the concatenation of the initial node features $\bff^{(0)}$ from Equation \eqref{eq:globalReadout}. Our experiments show that the addition of these features is beneficial for improved accuracy. Those findings are in line with other works that utilize $\bff^{(0)}$ such as \cite{klicpera2018combining,chen20simple} that obtain node-level representations. Here, we use $\bff^{(0)}$ to obtain better label features instead of local information.
\paragraph{The significance of channel expansion}
 We show the benefit of using the channel expansion in Equation \eqref{eq:SE} compared to using a standard expansion rate of $e = 1$ by inspecting the obtained accuracy with a variety of expansion rates. The results provided in Figure \ref{fig:expansion} show that the increase of $e$ helps to improve accuracy. \textcolor{black}{Specifically, when $e \geq 12$ we notice that the overall accuracy (across various datasets) does not further significantly improve, and therefore we choose $e=12$ in our experiments, also to balance between high accuracy and computational complexity}.

\paragraph{The influence of the cut-off value $\rm{r}$}
As depicted in Figure \ref{fig:cuttoff}, setting a low cut-off value (i.e., $\rm{r} < 1$) yields inferior results, as it promotes the first term in Equation \eqref{eq:globalLocalLoss}, that minimizes the distance between labels features and their corresponding nodes features. Analogously, setting $\rm{r} > 100$ will gravitate the network optimization towards the second term that maximizes the distance between node features and the features of labels that they do not correspond to. We found that $r=10$ yields the highest accuracy on average and we therefore use it throughout the rest of our experiments.

\paragraph{\textcolor{black}{The impact of the dropout probability $\rm{p}$}}
\textcolor{black}{We study the effect of the dropout value $p$ on the performance of GLGNN on the Cora, Citeseer, and Pubmed datasets. The results, reported in Figure \ref{fig:dropout} show that dropout is beneficial to improve accuracy.}

\paragraph{\textcolor{black}{The influence of the loss balancing coefficient $\gamma$}}
\textcolor{black}{We experiment with different values of $\gamma$ from Equation \eqref{eq:totalObjective}, and provide the results in Figure \ref{fig:gammaValue}. Our experiments show that including the proposed global-local loss in Equation \eqref{eq:totalObjective} improves the performance on the Cora, Citeseer and Pubmed datasets. Also, we found that as $\gamma$ is increased, the performance may be sub-optimal compared to using a small value of $\gamma$, as it balances the minimization of the cross entropy and global-local losses.}

\section{Conclusion}
\label{sec:conclusion}
In this paper, we propose GLGNN, a method to leverage global information for semi- and fully-supervised node classification,  \textcolor{black}{as well as graph classification.} \textcolor{black}{Our experimental results indicate that this information is beneficial to improve the performance of GNNs.}
Our method is based on learning global label features and fusing them with local node features. \textcolor{black}{Therefore, our method is suitable for classification tasks.} We show that it is possible to cluster the nodes in a way that enables improved classification accuracy and demonstrate that our method outperforms baseline models \textcolor{black}{such as GCN, GAT, and GCNII} by a large margin. \textcolor{black}{Furthermore, \textcolor{black}{as we show in our experiments,} GLGNN does not add significant computational costs, and can be potentially applied to other GNN architectures.}
Future research directions include the exploration of additional methods of global label information extraction and incorporation for further accuracy improvements \textcolor{black}{such as global graph structural properties}, \textcolor{black}{ as well as the use of pseudo-label learning for non-classification tasks, such as node property regression, where no classes or labels are available.}
\newline \\ 
\textbf{Data statement.} All datasets used in this paper are publicly available and are appropriately cited in the main manuscript.

\textbf{Funding disclosure.}
This research was supported by grant no. 2018209 from the United States - Israel
Binational Science Foundation (BSF), Jerusalem, Israel, and by the Israeli Council for Higher Education (CHE) via the Data Science Research Center. ME
is supported by Kreitman High-tech scholarship.

 \bibliographystyle{elsarticle-num} 
 \bibliography{bibli}





\end{document}